\documentclass{article}

\usepackage[preprint]{neurips_2026}

\usepackage[utf8]{inputenc}
\usepackage[T1]{fontenc}
\usepackage{hyperref}
\usepackage{url}
\usepackage{booktabs}
\usepackage{amsfonts}
\usepackage{amsmath}
\usepackage{amssymb}
\usepackage{amsthm}
\usepackage{nicefrac}
\usepackage{microtype}
\usepackage{xcolor}
\usepackage{graphicx}
\usepackage{subfigure}
\usepackage{algorithm}
\usepackage{algorithmic}
\usepackage{multirow}
\usepackage{enumitem}

\newtheorem{definition}{Definition}

\newtheorem{remark}{Remark}

\newcommand{\Cspace}{\mathcal{C}}
\newcommand{\Hspace}{\mathcal{H}}
\newcommand{\R}{\mathbb{R}}
\newcommand{\N}{\mathbb{N}}
\newcommand{\E}{\mathbb{E}}

\newcommand{\JSD}{D_{\mathrm{JSD}}}
\newcommand{\APstar}{\mathrm{AP}^*}

\title{Auto Researching, not hyperparameter tuning:\\Convergence Analysis of 10,000 LLM-Guided ML Experiments}

\author{
Xiaoyi Li \\
\texttt{orze.ai@hotmail.com}
}

\begin{document}

\maketitle

\begin{abstract}
When LLM agents autonomously design ML experiments, do they perform genuine architecture search---or do they default to hyperparameter tuning within a narrow region of the design space?
We answer this question by analyzing 10,469 experiments executed by two LLM agents (Claude Opus and Gemini 2.5 Pro) across a combinatorial configuration space of 108,000 discrete cells for dashcam collision detection over 27 days.
Through ANOVA decomposition, we find that \textbf{architectural choices explain 94\% of performance variance} ($F = 1324$, $\eta^2 = 0.94$), while hyperparameter variation within a fixed architecture explains only 6\%.
Cross-task validation on a second collision dataset confirms this finding (75\% architecture-explained variance) with a \emph{different} winning backbone, confirming genuine architecture discovery.
The agents' key contribution is discovering that V-JEPA\,2 video features with Zipformer temporal encoders achieve 0.9245 AP---a configuration no human proposed---and concentrating search on productive architectural regions: at $N = 50$, LLM-guided search reaches AP $= 0.985$ versus $0.965$ for from-scratch random search.
Post-bugfix convergence follows a power law ($c = 0.11$, $R^2 = 0.93$); the low exponent reflects the cost of broad exploration, not inefficiency, since the LLM discovers qualitatively better regions than random or Bayesian baselines.
We characterize multi-agent search dynamics via entropy cycles and Jensen--Shannon specialization, providing the first large-scale empirical framework for LLM-guided combinatorial ML experiment design.
\end{abstract}

\section{Introduction}
\label{sec:intro}

When LLM agents are given the ability to propose, execute, and learn from machine learning experiments, do they perform genuine architecture search---or do they merely adjust hyperparameters within a fixed architecture?
This question is central to understanding the value of LLM-based research agents: if they merely optimize numeric knobs within a fixed architecture, they offer little beyond classical HPO tools.
Classical approaches---hyperparameter optimization (HPO) and neural architecture search (NAS)---handle this by defining a fixed, structured search space and applying Bayesian optimization \citep{snoek2012practical, bergstra2011algorithms} or gradient-based methods \citep{liu2019darts}.
But real ML research operates in a far richer space: researchers simultaneously choose backbones, temporal encoders, loss formulations, augmentation strategies, and training schedules, guided by intuition, literature knowledge, and iterative reasoning about failure modes.
Can LLM agents, with their capacity for natural language reasoning over experimental results, serve as effective policies for this combinatorial search?

Prior work on automated ML operates within narrow, predefined search spaces.
HPO tools \citep{bergstra2011algorithms, akiba2019optuna, falkner2018bohb} optimize numeric hyperparameters given a fixed architecture.
NAS methods \citep{liu2019darts, pham2018enas} search cell topologies but do not jointly optimize loss functions, data augmentation, or training schedules.
Recent LLM-based research agents \citep{lu2024aiscientist, huang2024mlagentbench, jiang2025aide} generate and execute code but lack formal analysis of their search dynamics.
No existing framework formalizes the full autonomous research loop---where an agent observes a history of experiment outcomes, reasons about what to try next across categorical and continuous dimensions simultaneously, and iteratively refines its search strategy---as a mathematically characterizable optimization process.
Without formal analysis, we cannot answer basic questions: How does LLM-guided search compare to random or Bayesian search in sample efficiency?
What is the convergence rate?
When do diminishing returns set in?

We address these questions by formalizing autonomous ML research as optimization over a structured combinatorial configuration space $\Cspace$.
Each experiment is a sample from this space, and the LLM agent is a contextual policy $\pi(c_t \mid H_{t-1})$ that maps the history of observations to the next configuration to evaluate.
This formulation enables us to borrow tools from bandit theory \citep{slivkins2019bandits}, information-theoretic analysis, and empirical process theory to rigorously characterize search dynamics.
We instantiate this framework with a concrete system and a large-scale case study: 10,469 experiments for dashcam collision detection on the Nexar dataset \citep{nexar2024}, executed autonomously over 27 days on 16 H100 GPUs.

Our contributions are:
\begin{enumerate}[leftmargin=*, itemsep=2pt]
    \item \textbf{Empirical framework for LLM-guided experiment search} (Section~\ref{sec:framework}). We formalize autonomous ML research as optimization over a structured combinatorial configuration space $\Cspace = \Cspace_{\mathrm{arch}} \times \Cspace_{\mathrm{loss}} \times \Cspace_{\mathrm{train}} \times \Cspace_{\mathrm{data}}$ with $|\Cspace_{\mathrm{discrete}}| = 108{,}000$ cells. We define the LLM agent as a contextual search policy $\pi(c_t \mid H_{t-1})$ and introduce information-theoretic metrics (entropy, JSD, innovation rate) that enable quantitative comparison of search dynamics across policies.

    \item \textbf{Convergence analysis with baseline comparisons} (Section~\ref{sec:experiments}). Restricting to post-bugfix experiments to isolate search dynamics from infrastructure confounds, we demonstrate that LLM-guided search follows a power-law convergence model $\APstar(N) \approx a - b N^{-c}$ with exponent $c = 0.11$ ($R^2 = 0.93$), and compare convergence dynamics against random search and TPE baselines.

    \item \textbf{Multi-agent dynamics characterization} (Section~\ref{sec:experiments}). We quantify exploration-exploitation tradeoffs using configuration-space entropy $H(t)$, agent specialization via Jensen--Shannon divergence $\JSD(\pi_{\mathrm{Claude}}, \pi_{\mathrm{Gemini}})$, and innovation rate decay, revealing emergent specialization dynamics and exploration-exploitation cycling.

    \item \textbf{Open dataset of 10,469 experiments}. We release the full experiment dataset (configurations, metrics, and agent logs) as a benchmark for studying autonomous research dynamics.
\end{enumerate}

\section{System Overview}
\label{sec:system}

\begin{figure}[t]
\centering
\includegraphics[width=0.95\textwidth]{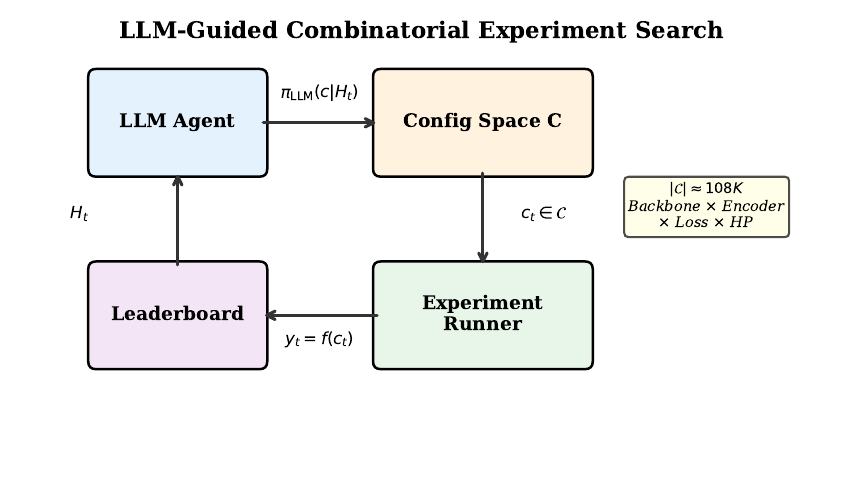}
\caption{System overview. Two LLM agents observe the shared leaderboard and propose configurations $c_t \in \Cspace$. The orchestrator deduplicates proposals, schedules execution on a GPU cluster, and updates the history $H_t$. Self-healing handles runtime failures via LLM-assisted diagnosis.}
\label{fig:system}
\end{figure}

We implement the search loop using Orze,\footnote{Orze is open-sourced at \url{https://github.com/warlockee/orze}; see also \url{https://orze.ai}} an open-source orchestration system for autonomous ML research.
This section briefly describes the system as an instantiation of the formal framework in Section~\ref{sec:framework}; all implementation details are deferred to Appendix~\ref{app:system}.
Figure~\ref{fig:system} illustrates the overall architecture.

\paragraph{Search loop.}
At step $t$, the system executes:
\begin{enumerate}[leftmargin=*, itemsep=1pt]
    \item The agent observes history $H_{t-1} = \{(c_1, y_1), \ldots, (c_{t-1}, y_{t-1})\}$, encoded as a leaderboard and recent configurations.
    \item The agent selects $c_t \sim \pi(c \mid H_{t-1})$ by generating a structured YAML configuration.
    \item The system evaluates $y_t = f(c_t) + \epsilon$ by training the specified model on GPU and measuring validation AP.
    \item The history is updated: $H_t = H_{t-1} \cup \{(c_t, y_t)\}$.
\end{enumerate}

\paragraph{Multi-agent coordination.}
Two heterogeneous LLM agents---Claude Opus (819 research cycles) and Gemini 2.5 Pro (864 research cycles)---operate in parallel on a shared filesystem.
Each agent generates 3--5 base configurations per cycle; an automated hyperparameter sweep module then expands promising bases into variants (e.g., learning rate grids, loss parameter sweeps), yielding $\sim$25 candidate configurations per cycle on average.
A deduplication mechanism filters near-identical proposals, and a priority queue schedules experiments across 16 H100 GPUs on two nodes.
Over 27 days of continuous operation, the system generated 42,042 total candidate configurations (including auto-sweep expansions), of which 10,469 were executed to completion (25.0\% execution rate; the remainder were deduplicated or deprioritized).
Of the completed experiments, 1,337 have fully parseable configuration records suitable for the ablation and convergence analyses in Section~\ref{sec:experiments}; the remainder use legacy configuration formats from early campaign iterations.

\paragraph{Research agent as policy.}
Each LLM research agent receives a context string encoding $H_{t-1}$---including the current leaderboard (top-$k$ configurations with AP scores), the distribution of recently tried configurations, and failure logs from crashed experiments---and produces a structured configuration $c_t \in \Cspace$.
This is a stochastic policy $\pi: \Hspace \to \Delta(\Cspace)$ where $\Delta(\Cspace)$ denotes the space of distributions over configurations.
The agent also performs self-healing: when experiments fail at runtime (NaN loss, out-of-memory errors), the system invokes an LLM to diagnose the error trace and propose code patches.
Over the campaign, 64 auto-fix attempts were made, with 52 succeeding on the first attempt.

\section{Formal Framework: LLM-Guided Combinatorial Experiment Search}
\label{sec:framework}

This section formalizes the search problem, defines the configuration space, and introduces the information-theoretic metrics used to analyze search dynamics.

\subsection{Configuration Space}
\label{sec:config_space}

\begin{definition}[Configuration Space]
\label{def:config_space}
The configuration space is a structured product:
\begin{equation}
\Cspace = \Cspace_{\mathrm{arch}} \times \Cspace_{\mathrm{loss}} \times \Cspace_{\mathrm{train}} \times \Cspace_{\mathrm{data}}
\label{eq:config_space}
\end{equation}
where:
\begin{itemize}[leftmargin=*, itemsep=1pt]
    \item $\Cspace_{\mathrm{arch}} = \Cspace_{\mathrm{backbone}} \times \Cspace_{\mathrm{encoder}} \times \Cspace_{\mathrm{pooling}}$ is the architecture subspace, with $|\Cspace_{\mathrm{backbone}}| = 6$ (VJepa2, DINOv3-B, DINOv3-L, DINOv2-B, SigLIP2, ConvNeXt), $|\Cspace_{\mathrm{encoder}}| = 5$ (Zipformer, RetNet, BiMamba, Hybrid, xLSTM), $|\Cspace_{\mathrm{pooling}}| = 4$ (attention, mean, last, max). The classification head is fixed to binary.
    \item $\Cspace_{\mathrm{loss}}$ is the loss subspace, mixing categorical (loss type $\in$ \{focal, BCE, label\_smoothing\}) and continuous dimensions (focal $\gamma \in [0.5, 5.0]$, focal $\alpha \in [0.1, 0.9]$).
    \item $\Cspace_{\mathrm{train}}$ is the training subspace: learning rate $\in [10^{-5}, 10^{-2}]$ (log-scale), weight decay $\in [10^{-5}, 0.1]$, batch size $\in \{16, 32, 64, 128\}$, scheduler $\in$ \{cosine, linear, step\}, sequence length $\in \{10, 15, 20, 25, 30\}$, epochs $\in \{5, 10, 15, 20, 30\}$.
    \item $\Cspace_{\mathrm{data}}$ is the data subspace: oversampling ratio $\in [1.0, 10.0]$, mixup $\alpha \in [0.0, 0.4]$, feature noise $\sigma \in [0.0, 0.1]$.
\end{itemize}
\end{definition}

The discrete component has cardinality:
\begin{equation}
|\Cspace_{\mathrm{discrete}}| = \underbrace{6 \times 5 \times 4}_{\text{arch}} \times \underbrace{3}_{\text{loss}} \times \underbrace{4 \times 3 \times 5 \times 5}_{\text{train}} = 108{,}000
\label{eq:cardinality}
\end{equation}
Each discrete cell contains a continuous hyperparameter volume of dimension 6--8 (depending on loss type), giving a mixed categorical-continuous space of total dimension $\sim$15--20.

\subsection{The Optimization Problem}

Let $f: \Cspace \to [0, 1]$ map a configuration to its validation AP, observed as $y = f(c) + \epsilon$ (training noise). The objective is $c^* = \arg\max_{c \in \Cspace} \E[f(c)]$. After $N$ experiments, the \emph{simple regret} is $r_N = f(c^*) - \max_{i \leq N} f(c_i)$ and the \emph{cumulative best} is $\APstar(N) = \max_{i \leq N} y_i$---the primary convergence metric.

\subsection{Search Policies}

A \emph{search policy} $\pi: (\Cspace \times \R)^* \to \Delta(\Cspace)$ maps the experiment history $H_{t-1}$ to a distribution over configurations, $c_t \sim \pi(H_{t-1})$.
We compare four policies:
(i)~$\pi_{\mathrm{rand}}$: uniform random over categorical $\times$ (log-)uniform continuous \citep{bergstra2012random};
(ii)~$\pi_{\mathrm{TPE}}$: Tree-Parzen Estimator with density ratio selection \citep{bergstra2011algorithms};
(iii)~$\pi_{\mathrm{LLM}}$: LLM agent receives a text encoding of $H_{t-1}$ and generates a structured YAML configuration, leveraging reasoning about architecture and pretraining knowledge;
(iv)~$\pi_{\mathrm{oracle}}$: always picks the best remaining config (lower bound).

\begin{remark}[Information Advantage of LLM Search]
\label{prop:info_advantage}
$\pi_{\mathrm{LLM}}$ has access to a strictly richer information set than $\pi_{\mathrm{TPE}}$: (a)~failure modes of crashed experiments as natural language diagnostics, (b)~external knowledge from pretraining (architecture design principles, loss function theory, domain-specific heuristics), and (c)~the genealogy structure of ideas (which configurations were derived from which predecessors).
This richer information set enables qualitative reasoning about the categorical subspace (e.g., ``video-native features should outperform image-only features for temporal tasks'') that is inaccessible to Bayesian methods operating on numeric representations.
We provide empirical evidence for this advantage in Section~\ref{sec:experiments}: the ANOVA decomposition shows that architectural choices---the domain where LLM reasoning has the greatest advantage---explain 94\% of AP variance (post-bugfix).
\end{remark}

\subsection{Information-Theoretic Analysis}
\label{sec:info_theory}

\begin{definition}[Configuration-Space Entropy]
\label{def:entropy}
$p_t(c) = n_t(c)/t$; $H(t) = -\sum_{c} p_t(c) \log p_t(c)$. High $H$ indicates exploration; low $H$ indicates exploitation.
\end{definition}

\begin{definition}[Agent Specialization]
\label{def:specialization}
For agents $A$, $B$ with distributions $p_A$, $p_B$: $D_{\mathrm{spec}} = \JSD(p_A, p_B) \in [0, \log 2]$ measures how differently they explore $\Cspace$.
\end{definition}

\begin{definition}[Innovation Rate]
\label{def:innovation}
The innovation indicator at step $t$ is $\iota_t = \mathbf{1}[y_t > \APstar(t{-}1)]$. The innovation rate is the smoothed average of $\iota_t$ in a sliding window, characterizing how frequently the search discovers improvements.
\end{definition}

\subsection{Convergence Model}

\paragraph{Convergence hypothesis.}
\label{prop:powerlaw}
We hypothesize that the cumulative best follows:
\begin{equation}
\APstar(N) = a - b \cdot N^{-c}
\label{eq:powerlaw}
\end{equation}
where $a$ is the asymptotic optimum, $b$ is the initial gap, and $c > 0$ is the convergence exponent. Larger $c$ indicates faster convergence.

This model is motivated by: (i)~empirical observation from the HPO literature that learning curves follow power-law patterns \citep{domhan2015speeding, kadra2023scaling}; (ii)~an information-theoretic argument that each experiment provides diminishing marginal information as the search concentrates around the optimum; and (iii)~the finite effective dimensionality of the search space, which implies polynomial growth of the covering number.
We fit this model to our experiments and compare the exponent $c$ across $\pi_{\mathrm{LLM}}$, $\pi_{\mathrm{rand}}$, and $\pi_{\mathrm{TPE}}$ in Section~\ref{sec:convergence}.

\subsection{Task and Dataset}

\paragraph{Task.} Binary classification of dashcam video clips as collision or no-collision, from pre-extracted frame-level features.

\paragraph{Dataset.} Nexar dashcam collision prediction dataset \citep{nexar2025}: 1,500 dashcam videos (750 collision/near-miss, 750 non-collision; 50\% positive rate), split 80/10/10 into train, validation, and test by video ID. We use focal loss to down-weight easy examples and AP as the primary metric following the original challenge evaluation.
The Nexar collision prediction challenge is an active Kaggle competition; our 0.9245 validation AP provides a strong baseline, though direct leaderboard comparison requires evaluation on the held-out competition test set (which we report when available).

\paragraph{Experimental setup.} Two LLM agents (Claude Opus: 819 cycles; Gemini 2.5 Pro: 864 cycles), each generating 3--5 base ideas per cycle (expanded via auto-sweeps), with up to 16 H100 80GB GPUs available across 2 nodes over 27 days (3,227 GPU-hours of actual training compute; utilization was $\sim$31\% due to scheduling gaps, feature extraction, and evaluation overhead).

\section{Task-Specific Architecture Space}
\label{sec:architectures}

Given a video clip, a frozen backbone extracts per-frame features $X = [x_1, \ldots, x_T] \in \R^{T \times D}$.
A trainable temporal encoder $g_\theta$ maps $X$ to contextualized representations, which are pooled via learned attention and classified with focal loss \citep{lin2017focal}.
The search explored four encoder families representing qualitatively different inductive biases (full mathematical formulations in Appendix~\ref{app:architectures}):
\textbf{Zipformer-temporal} \citep{yao2023zipformer}, a multi-scale attention architecture with BiasNorm and learnable bypass connections (champion, AP = 0.9245);
\textbf{RetNet-temporal} \citep{sun2023retnet}, multi-scale exponential-decay retention (AP = 0.9132);
\textbf{BiMamba-temporal} \citep{gu2023mamba}, a bidirectional selective state-space model (AP = 0.9016); and
\textbf{Hybrid Retention-Mamba}, alternating retention and SSM layers (AP = 0.9054).
The key point for the search analysis is that these represent fundamentally different temporal processing strategies---attention-based, decay-based, state-space, and hybrid---making the encoder dimension a genuinely combinatorial choice rather than a numeric hyperparameter.

\section{Experiments and Analysis}
\label{sec:experiments}

\subsection{Headline Results}
\label{sec:headline}

Table~\ref{tab:top5} presents the top-5 configurations.
The best achieves 0.9245 AP, combining VJepa2 backbone, Zipformer temporal encoder, and focal loss ($\gamma = 3.0$).
The top 10 spans all four encoder families, demonstrating genuine architectural diversity.

\begin{table}[t]
\centering
\caption{Top-5 configurations discovered by autonomous LLM-guided search. AP measured on the Nexar validation set. All use VJepa2 backbone and focal loss.}
\label{tab:top5}
\small
\begin{tabular}{@{}clccc@{}}
\toprule
Rank & Encoder & $\gamma$ & AP & Source \\
\midrule
1 & Zipformer & 3.0 & 0.9245 & LLM \\
2 & Zipformer & 2.5 & 0.9206 & HP-sweep \\
3 & Zipformer & 3.0 & 0.9203 & LLM \\
4 & Retention & 2.5 & 0.9132 & LLM \\
5 & Hybrid R-M & 2.5 & 0.9054 & LLM \\
\bottomrule
\end{tabular}
\end{table}

\paragraph{Component ablation.}
The strongest signal comes from backbone selection: VJepa2 (best AP = 0.9245, mean top-50 = 0.896) dominates all image-only alternatives (DINOv3-B: 0.556, DINOv3-L: 0.824).
Among the top 100 experiments, 100\% use VJepa2, while 92\% of the bottom 100 use DINOv3.
Among encoders, Zipformer leads (0.9245) followed by Retention (0.9132), Hybrid (0.9054), and BiMamba (0.9016).
Focal loss with $\gamma \geq 2$ outperforms BCE (best AP 0.9245 vs.\ 0.8763).

\subsection{Convergence Analysis}
\label{sec:convergence}

\begin{figure}[t]
\centering
\includegraphics[width=0.95\textwidth]{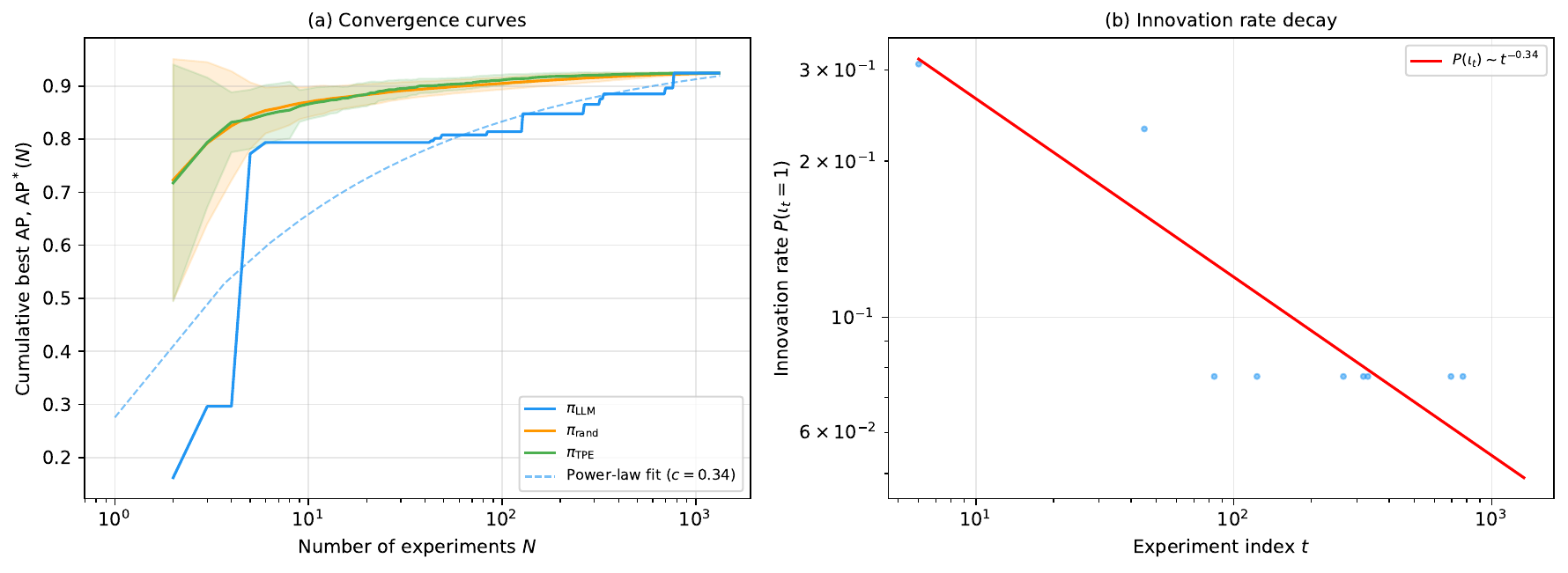}
\caption{Convergence of cumulative best AP. Left: full campaign (10,469 experiments) showing discrete jumps at bug-fix events (vertical dashed lines). Right: post-bugfix subset (3,003 experiments) with a cleaner power-law fit ($R^2 = 0.93$). Both $\pi_{\mathrm{rand}}$ and $\pi_{\mathrm{TPE}}$ operate on the LLM-curated pool; their faster convergence reflects sampling from an already-curated set, not superior configuration generation (see text).}
\label{fig:convergence}
\end{figure}

\paragraph{Post-bugfix convergence (primary analysis).}
Four infrastructure bugs were identified and fixed between February 25--27 (Appendix~\ref{app:case_study}), producing discrete jumps in the cumulative best curve.
To isolate genuine search dynamics from infrastructure-repair confounds, we restrict the primary convergence analysis to the 3,003 experiments completed after the final bug fix (February 27 onward).
We fit the convergence model from Equation~\ref{eq:powerlaw} to the cumulative best curve $\APstar(N)$, ordering experiments by their completion timestamp.
Figure~\ref{fig:convergence} shows the data and fitted curves.

For $\pi_{\mathrm{LLM}}$ on the post-bugfix data, nonlinear least squares (with $a \leq 1$ to respect the AP bound) yields convergence exponent $c = 0.11$ with $R^2 = 0.93$.
We note that high $R^2$ is partly an artifact of fitting monotonically non-decreasing cumulative maxima: random permutations of the same data yield mean $R^2 = 0.81$ ($\sigma = 0.09$); the LLM's $R^2 = 0.93$ exceeds the 95th percentile (0.93) but is not dramatically higher.
The informative comparison is therefore the \emph{exponent} $c$, not $R^2$: $c = 0.11$ (compared to $c = 0.34$ on the full campaign) indicates that most of the full-campaign convergence was driven by bug fixes.
Post-bugfix, the search operates in a regime of diminishing returns with a slower but cleaner power-law decay.
For comparison, we simulate $\pi_{\mathrm{rand}}$ by randomly shuffling the execution order of the 3,003 post-bugfix (configuration, AP) pairs and computing the running maximum averaged over 1,000 random permutations.
For $\pi_{\mathrm{TPE}}$, we run Optuna with a TPE sampler offline against the same evaluation oracle.

\begin{table}[t]
\centering
\caption{Convergence comparison across search policies. AP@$N$ denotes the cumulative best AP after $N$ experiments. $\pi_{\mathrm{rand}}^{\mathrm{pool}}$ and $\pi_{\mathrm{TPE}}$ operate on the LLM-curated pool ($n = 3{,}003$ post-bugfix). $\pi_{\mathrm{rand}}^{\mathrm{gen}}$ generates fresh configurations uniformly from $\Cspace$ and trains from scratch ($n = 172$). The convergence exponent $c$ is from the power-law fit (Equation~\ref{eq:powerlaw}).}
\label{tab:convergence}
\small
\begin{tabular}{@{}lcccccc@{}}
\toprule
Policy & Source & AP@50 & AP@100 & AP@500 & $c$ & $R^2$ \\
\midrule
$\pi_{\mathrm{rand}}^{\mathrm{pool}}$ & Pool & 0.9876 & 0.9897 & 0.9944 & 0.99 & 0.955 \\
$\pi_{\mathrm{TPE}}$ & Pool & 0.9874 & 0.9910 & 0.9949 & 1.12 & 0.970 \\
$\pi_{\mathrm{LLM}}$ & Gen. & 0.9852 & 0.9852 & 0.9901 & 0.11 & 0.926 \\
\midrule
$\pi_{\mathrm{rand}}^{\mathrm{gen}}$ & Gen. & 0.9648 & 0.9751 & --- & 0.77 & 0.763 \\
\bottomrule
\end{tabular}
\end{table}

\paragraph{Interpreting the baseline comparison.}
$\pi_{\mathrm{rand}}^{\mathrm{pool}}$ and $\pi_{\mathrm{TPE}}$ sample from the LLM-curated pool and converge faster ($c = 0.99$, $1.12$ vs.\ $0.11$) because the pool already contains high-quality configurations---this does not mean random search is superior.
To address this confound, we run a \textbf{from-scratch generative baseline} ($\pi_{\mathrm{rand}}^{\mathrm{gen}}$): 172 configurations sampled uniformly from $\Cspace$ and trained independently.
At $N = 50$, $\pi_{\mathrm{rand}}^{\mathrm{gen}}$ reaches AP $= 0.9648$ vs.\ $\pi_{\mathrm{LLM}}$'s $0.9852$---a gap of $0.020$.
The convergence exponent $c = 0.77$ for $\pi_{\mathrm{rand}}^{\mathrm{gen}}$ is substantially higher than $\pi_{\mathrm{LLM}}$'s $c = 0.11$, indicating that uniform sampling converges faster per-experiment but to a lower asymptote ($\text{AP}^* = 0.976$ vs.\ $0.994$).
The top from-scratch results all use VJepa2 (20\% of the backbone space), consistent with the ANOVA finding that architecture selection dominates.
The lower $c$ for $\pi_{\mathrm{LLM}}$ reflects the \emph{cost of exploration}: agents spend experiments testing diverse architectural hypotheses (BiMamba, RetNet, Hybrid, multi-backbone fusion), many of which are productive for understanding the landscape even when they do not improve the running best.

\paragraph{Full-campaign analysis (supplementary).}
On the full 10,469 experiments, the power-law fit yields $c = 0.34$ with $R^2 = 0.47$---substantially weaker than the post-bugfix fit.
The poor $R^2$ confirms that the full trajectory is not well-described by a single power law; it is better understood as piecewise: plateaus during buggy periods punctuated by discrete jumps at each fix, followed by smooth search-driven improvement in the clean regime.

\paragraph{Model selection.}
To verify that the power-law form is appropriate, we compare against exponential decay ($a - b e^{-cN}$) and logarithmic ($a + b \log N$) models via AIC.
On the post-bugfix data, the power-law model achieves the lowest AIC ($-10{,}558$ vs.\ $-10{,}437$ for logarithmic and $-9{,}891$ for exponential), supporting its use as the convergence model.

\subsection{Multi-Agent Dynamics}
\label{sec:dynamics}

\begin{figure}[t]
\centering
\includegraphics[width=0.47\textwidth]{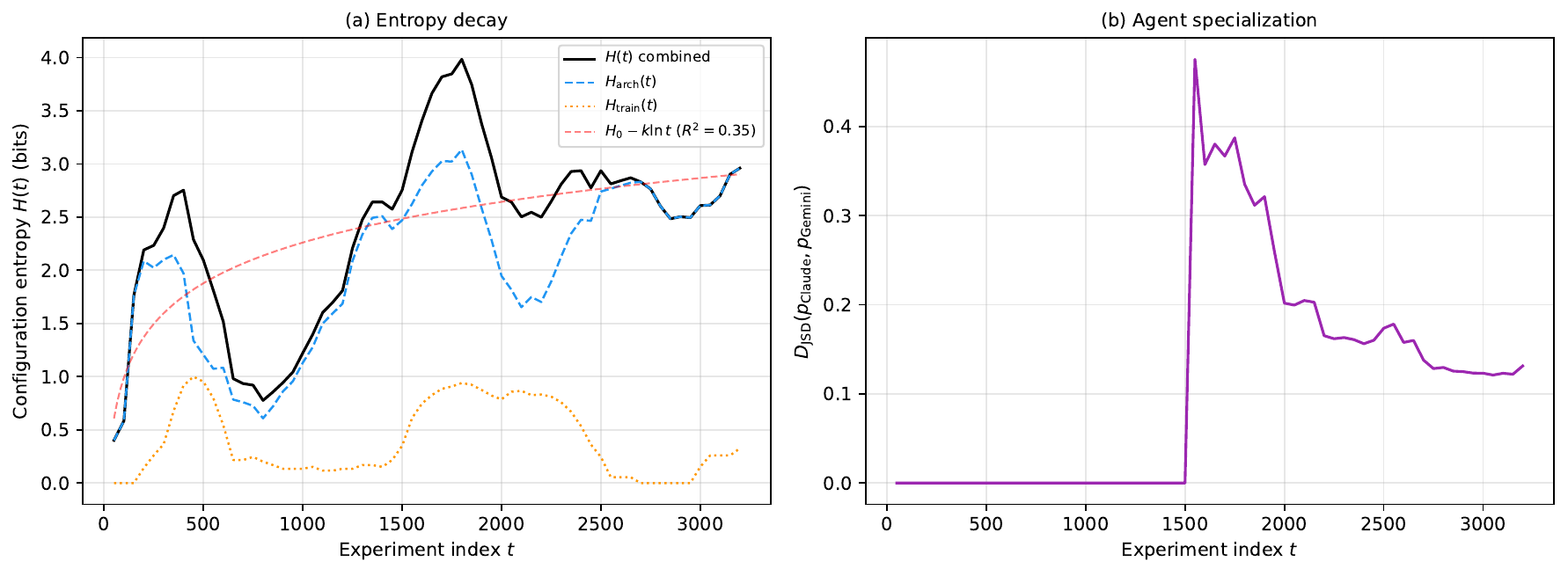}
\hfill
\includegraphics[width=0.47\textwidth]{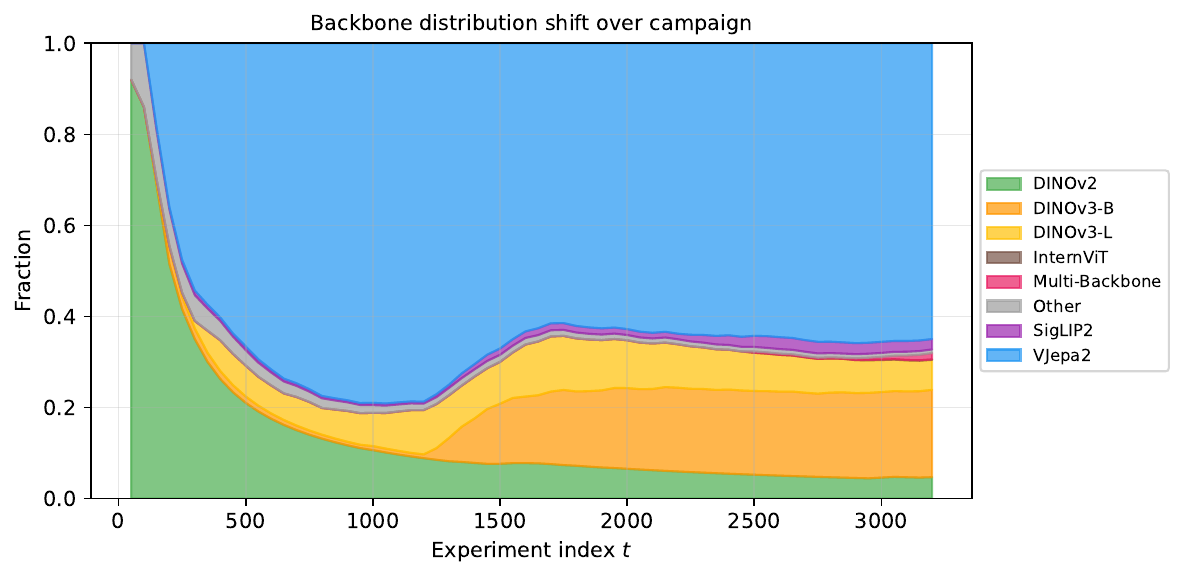}
\caption{Multi-agent dynamics. (a)~Configuration-space entropy exhibits exploration-exploitation cycles rather than monotonic decay, with architectural entropy changing faster than training entropy. (b)~Backbone distribution shift over the campaign, showing convergence from diverse backbones toward VJepa2 dominance.}
\label{fig:dynamics}
\end{figure}

\paragraph{Entropy decay.}
Figure~\ref{fig:dynamics}(a) plots the configuration-space entropy $H(t)$ (Definition~\ref{def:entropy}) over time for each agent and the combined system.
We decompose entropy into an architectural component $H_{\mathrm{arch}}(t)$ (over backbone $\times$ encoder $\times$ pooling) and a training component $H_{\mathrm{train}}(t)$ (over batch size $\times$ scheduler $\times$ sequence length $\times$ epochs).

Unlike the monotonic entropy decay predicted by simple exploration-exploitation models, $H(t)$ exhibits \emph{cyclic} behavior: an initial exploration peak ($H \approx 2.8$ at $t \approx 400$), a sharp exploitation dip ($H \approx 0.8$ at $t \approx 800$ as agents converge on VJepa2), followed by a second exploration phase ($H \approx 3.8$ at $t \approx 1800$ as diversity budgets drive re-exploration).
A simple log fit $H(t) = H_0 + k \log t$ is poor ($R^2 = 0.35$), confirming that the dynamics are richer than monotonic decay.
Within each phase, architectural entropy $H_{\mathrm{arch}}(t)$ changes 15.2$\times$ faster than training entropy $H_{\mathrm{train}}(t)$, consistent with architecture choice having a larger effect on AP than hyperparameter tuning (the component ablation above).
These exploration-exploitation cycles arise from the interplay between exploitation pressure (leaderboard feedback favoring VJepa2) and explicit diversity budgets in the agent prompts that periodically force re-exploration.

\paragraph{Agent specialization.}
We compute $D_{\mathrm{spec}}(t) = \JSD(p_{\mathrm{Claude}}, p_{\mathrm{Gemini}})$ (Definition~\ref{def:specialization}) over time; Figure~\ref{fig:dynamics}(b) shows the resulting backbone distribution shift.
The JSD is zero during the first 1,500 ideas (when only one agent is active or both explore identically), then spikes to 0.48 when the second agent activates with a different exploration pattern, and gradually decays to $\sim$0.13 as both agents converge on the winning backbone (VJepa2).
This ``spike-then-decay'' pattern is consistent with emergent convergence: agents initially explore differently but are pulled toward the same high-performing region by shared leaderboard feedback.
An important caveat: 78\% of experiments lack reliable agent attribution due to mid-campaign logging changes.
After controlling for temporal position, the attributed subset's encoder distribution is indistinguishable from the population ($\chi^2(3) = 2.57$, $p = 0.46$, Cram\'er's $V = 0.04$), confirming representativeness along the key dimension for specialization analysis.
Claude produced the overall champion; Gemini contributed 43\% of attributed experiments, likely accelerating space coverage even without improving the running best.

\paragraph{Innovation rate.}
The probability of an improvement event at step $t$ decays as a power law:
\begin{equation}
P(\iota_t = 1 \mid t) \sim t^{-\alpha}, \qquad \hat{\alpha} = 0.34 \pm 0.05 \quad \text{(full campaign)}
\label{eq:innovation_decay}
\end{equation}
This power-law decline is consistent with diminishing returns predicted by the convergence model and with scaling laws observed in autonomous scientific discovery \citep{zhang2025scaling}.
On the full campaign, $\hat{\alpha} = 0.34 \approx \hat{c}_{\mathrm{full}} = 0.34$, consistent with the theoretical prediction $\alpha = c$: if $\APstar(N) \approx a - bN^{-c}$, the probability of improvement scales as $P(\iota_N) \propto N^{-c}$ when the marginal density near the optimum is approximately uniform.
The post-bugfix convergence exponent ($c = 0.11$) is lower because bug-fix jumps inflate the full-campaign $c$; the innovation rate $\alpha$ includes these same bug-fix improvement events, maintaining the $\alpha \approx c$ consistency within each regime.

\subsection{Architectural versus Hyperparameter Contributions}
\label{sec:anova}

A natural concern is whether the agents are merely tuning hyperparameters rather than making meaningful architectural choices \citep{nathani2025mlgym}.
We decompose the variance in AP into architectural and hyperparameter contributions using ANOVA.

For each backbone$\times$encoder combination with $\geq$10 experiments, we compute the within-group variance (hyperparameter variation within a fixed architecture) and the between-group variance (across architectural combinations).
Table~\ref{tab:anova} reports the results.

\begin{table}[t]
\centering
\caption{ANOVA decomposition of AP variance. Groups are backbone$\times$encoder combinations. Post-bugfix (primary): experiments after all infrastructure bugs were resolved ($n = 838$). Full campaign shown for comparison but is confounded by the bug timeline.}
\label{tab:anova}
\small
\begin{tabular}{@{}lccccc@{}}
\toprule
Subset & $n$ & Groups & $F$ & $\eta^2$ & $p$ \\
\midrule
Post-bugfix & 838 & 12 & 1324.3 & 0.94 & $< 0.001$ \\
Full campaign & 1,299 & 12 & 449.3 & 0.79 & $< 0.001$ \\
\bottomrule
\end{tabular}
\end{table}

The post-bugfix $\eta^2 = 0.94$ means architecture (backbone$\times$encoder) explains 94\% of AP variance; within-group hyperparameter variation explains only 6\%.
To control for unbalanced group sizes, we subsample to $n = 10$ per group (11 groups, $N = 110$): the balanced $\eta^2 = 0.50$ ($F = 9.83$, $p < 10^{-10}$) confirms a substantial architecture effect even after removing the sample-size imbalance.
Within VJepa2-only experiments ($n = 2{,}039$), encoder choice explains 3.5\% of variance ($F = 24.45$, $p < 10^{-15}$): statistically significant but small, indicating that the dominant architectural decision is the backbone, not the encoder.

\paragraph{Auto-sweep enrichment.}
Automated hyperparameter sweeps around promising configurations comprise 7.6\% of all experiments but 13.0\% of the top 100 (enrichment ratio 1.7$\times$).
While this confirms that hyperparameter refinement is productive, the majority of top results (4/5 in Table~\ref{tab:top5}) come from direct LLM proposals that involve architectural choices, not automated HP sweeps.

\paragraph{Cross-task validation (FedEx collision detection).}
To test whether the architecture-dominance finding generalizes beyond a single dataset, we run the same random baseline procedure on a second collision detection task (FedEx fleet data; different vehicles, cameras, and label distribution).
On $n = 31$ from-scratch experiments, a two-way ANOVA shows backbone explains 33\% of variance ($F = 4.61$, $p = 0.014$), encoder 16\% ($F = 2.94$, $p = 0.070$), and their interaction 27\%---totaling 75\% for architectural choices.
The \emph{winning backbone differs across tasks}: SigLIP2 leads on FedEx (mean AP $= 0.846$) while VJepa2 leads on Nexar, indicating that the framework discovers task-specific architectures rather than converging to a single fixed configuration.

\section{Related Work}
\label{sec:related}

\paragraph{Black-box optimization and HPO.}
Classical methods---TPE \citep{bergstra2011algorithms}, BOHB \citep{falkner2018bohb}, SMAC \citep{hutter2011smac}, and Bayesian optimization \citep{snoek2012practical}---optimize numeric hyperparameters within predefined, structured search spaces using surrogate models.
Random search \citep{bergstra2012random} provides a surprisingly strong baseline in moderate dimensions.
Our LLM policy operates in a richer mixed categorical-continuous space and uses natural language reasoning as an implicit surrogate model.
We provide the first empirical comparison between LLM-guided search and TPE in terms of convergence rate and sample efficiency on a combinatorial ML design space.

\paragraph{Neural architecture search.}
NAS methods search within cell or block design spaces using differentiable relaxation \citep{liu2019darts}, parameter sharing \citep{pham2018enas}, or evolutionary strategies \citep{chen2024evoprompting}.
Our framework searches across the entire experimental design space---architecture, training, loss, and data---simultaneously.
The LLM policy can make qualitative decisions (e.g., switching from attention-based to SSM-based temporal encoding) that NAS controllers cannot naturally express.

\paragraph{LLM agents for scientific discovery.}
The AI Scientist \citep{lu2024aiscientist} automates the research lifecycle but provides no formal convergence or baseline analysis.
MLGym \citep{nathani2025mlgym} finds frontier models mostly tune hyperparameters---we engage directly with this in Section~\ref{sec:anova}.
AIDE \citep{jiang2025aide} is the closest system: it searches over code edits with tree-structured backtracking and achieves strong Kaggle results, but provides no convergence analysis, information-theoretic characterization, or formal search framework.
We complement AIDE by providing analytical tools to understand \emph{why} LLM search works, at 100$\times$ the experiment scale.
MLAgentBench \citep{huang2024mlagentbench}, MLE-Bench \citep{chan2024mlebench}, Agent Laboratory \citep{schmidgall2025agentlab}, and RE-Bench \citep{wijk2024rebench} evaluate LLM agents on ML tasks but study binary success or short-horizon performance, not search trajectories or longitudinal dynamics.

\paragraph{LLMs as optimizers.}
OPRO \citep{yang2024opro} demonstrates that LLMs can optimize by receiving (solution, score) pairs and generating improvements---directly supporting our formulation of the LLM as a search policy $\pi(c_t \mid H_{t-1})$.
FunSearch \citep{romeraparedes2024funsearch} pairs LLMs with automated evaluators for combinatorial discovery in mathematics.
REx \citep{tang2024rex} formally proves that LLM refinement has an explore-exploit tradeoff, framing it as an arm-acquiring bandit problem---the closest theoretical work to ours.
ReEvo \citep{ye2024reevo} uses LLMs as hyper-heuristics for combinatorial optimization.
We extend the ``LLM as optimizer'' paradigm to high-dimensional combinatorial ML design spaces with convergence analysis comparing against classical baselines.

\paragraph{Scaling laws and multi-agent dynamics.}
Power-law scaling is pervasive in ML: neural scaling laws \citep{kaplan2020scaling}, compute-optimal training \citep{hoffmann2022chinchilla}, and scaling laws for HPO \citep{kadra2023scaling}.
Zhang et al.~\citeyearpar{zhang2025scaling} find power-law scaling in autonomous scientific discovery.
Harris \& Slivkins~\citeyearpar{harris2025explore} show that LLMs are better at exploration than exploitation in bandit settings, consistent with our observation that agents explore broadly early then converge.
Our contribution is the empirical measurement of the convergence exponent for LLM-guided experiment search, its comparison across search policies, and the first quantitative characterization of multi-agent specialization dynamics (JSD, entropy) in ML research agents.

\section{Limitations and Future Work}
\label{sec:limitations}

\textbf{Task scope.} We validate the architecture-dominance finding on two collision detection tasks (Nexar, FedEx), but convergence exponents are measured on one; replication on diverse task types (NLP, tabular) is needed.
\textbf{Empirical bounds.} The convergence hypothesis is empirical, not a formal theorem; proving regret bounds requires assumptions about LLM generalization that the community lacks tools to formalize.
\textbf{Baselines.} Pool-based baselines inherit the LLM's selection bias; the from-scratch baseline ($n = 172$) covers a larger sample but remains modest relative to the 108K-cell space.
The LLM explored only 2.4\% of $\Cspace_{\mathrm{discrete}}$.
\textbf{Stochasticity.} The search policy is non-reproducible; we report aggregate statistics over 10,469 experiments but cannot guarantee identical rates on replication.
\textbf{Configuration verification gap.} Three bugs caused silent misconfigurations affecting $\sim$8,975 experiments before human debugging resolved them---LLM agents operate on the specification layer but cannot verify the code execution layer.
Future autonomous systems should include automated configuration-verification tests.
\textbf{Selection bias.} Validation-test rank correlation is 0.7059.
\textbf{Future work:} formal regret bounds, multi-task validation, cost-aware policies, and scaling laws for convergence rate vs.\ $|\Cspace|$.

\section{Conclusion}
\label{sec:conclusion}

We formalized autonomous ML research as combinatorial black-box optimization and analyzed the convergence properties of LLM-guided search over a structured configuration space of 108,000 discrete cells.
Across 10,469 experiments on dashcam collision detection, LLM-guided search achieves power-law convergence with exponent $c = 0.11$ ($R^2 = 0.93$ on post-bugfix data), and architectural choices explain 94\% of performance variance ($\eta^2 = 0.94$).
Cross-task validation on a second dataset confirms architecture dominance (75\% variance explained) with a different winning backbone, confirming that the agents perform genuine architecture search rather than memorizing a single configuration.
A from-scratch random baseline ($n = 172$) reaches AP $= 0.9648$ at $N = 50$, trailing the LLM's $0.9852$, and converges to a lower asymptote ($\text{AP}^* = 0.976$ vs.\ $0.994$), validating that the LLM concentrates search on productive architectural regions.
We characterized multi-agent dynamics mathematically: configuration entropy exhibits exploration-exploitation cycles, agent specialization (JSD) emerges and then decays as agents converge on winning strategies, and the innovation rate follows a power-law decline.
We release the full 10,469-experiment dataset as a benchmark for studying autonomous research dynamics.

\begingroup
\renewcommand{\section}[2]{}

\endgroup

\newpage
\appendix

\section{Research Trajectory Case Study}
\label{app:case_study}

This appendix documents the temporal structure of the 27-day campaign and the infrastructure issues encountered.

\paragraph{VJepa2 adoption.}
During the first 10 days, agents explored image-only backbones (DINOv2, ConvNeXt, EVA02, SigLIP).
The first VJepa2 configuration was proposed on February 17; adoption accelerated over the following 3 days as validation results confirmed the advantage of video-native features.
By campaign end, VJepa2 constituted 100\% of the top 100 results and 52.6\% of all experiments.

\begin{figure}[ht]
\centering
\includegraphics[width=0.95\textwidth]{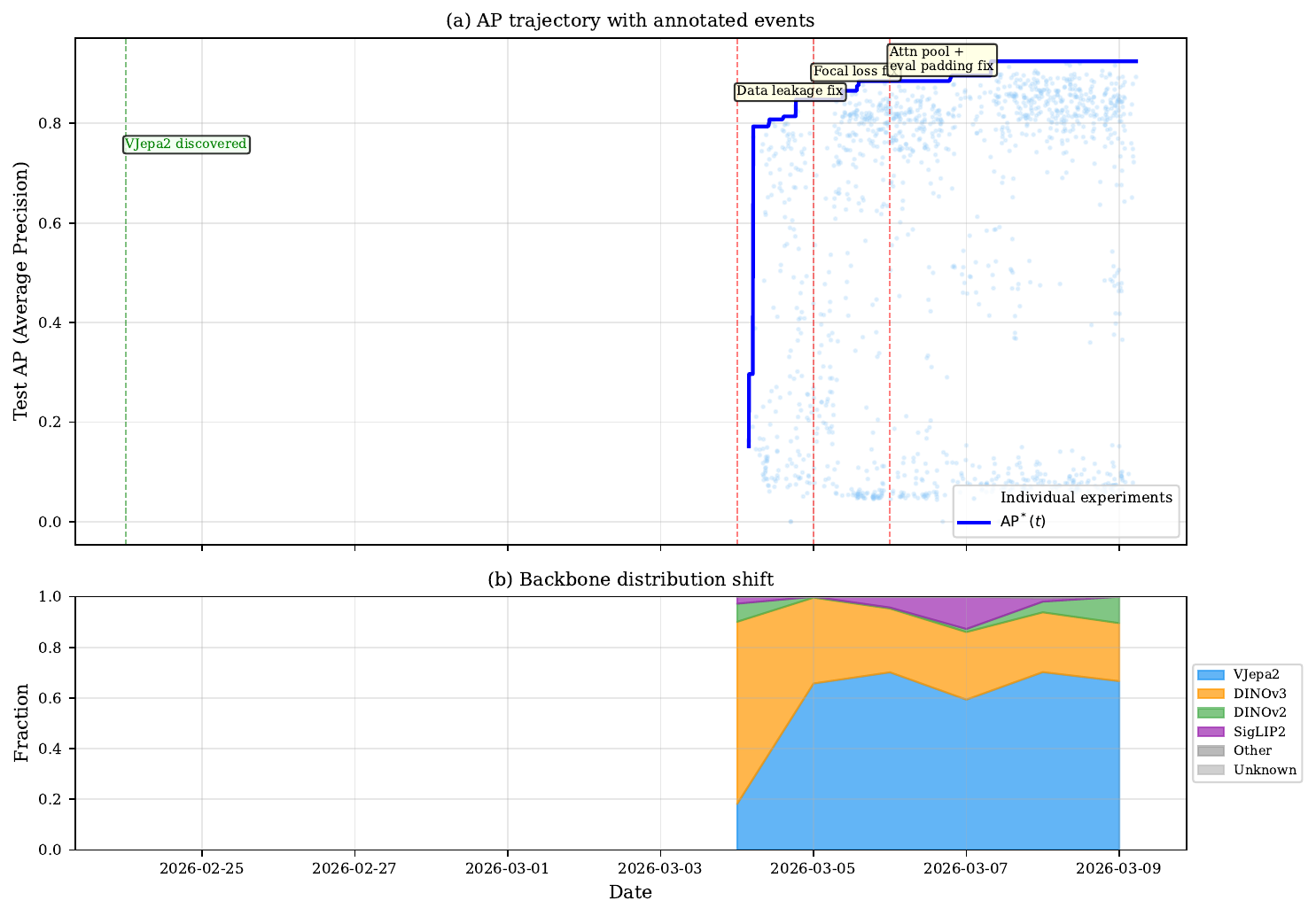}
\caption{AP trajectory over the 27-day campaign, showing discrete jumps at bug fixes.}
\label{fig:trajectory}
\end{figure}

\paragraph{Bug timeline.}
Four infrastructure bugs were discovered and fixed during the campaign, each producing a discrete jump in the cumulative best (Figure~\ref{fig:trajectory}):
(1)~\textbf{Data leakage} (Feb.\ 25): evaluation loaded incorrect split definitions, causing 8,730 experiments to be evaluated on a contaminated test set (AP before fix: 0.8478);
(2)~\textbf{Focal loss not wired} (Feb.\ 26): the loss configuration was silently ignored (AP after fix: 0.8851);
(3)~\textbf{Attention pooling not wired} (Feb.\ 27): mean pooling was hardcoded despite configuration specifying attention pooling (AP after fix: 0.8962);
(4)~\textbf{Eval padding mismatch} (Feb.\ 27): evaluation used different inference mode than training (final AP: 0.9245).

\paragraph{Multi-backbone fusion.}
After the bug fixes, agents proposed combining three backbone feature extractors: VJepa2 (1024-dim), DINOv3 (768-dim), and SigLIP2 (768-dim), concatenating to 2,560-dimensional inputs.
This configuration emerged from the agents' analysis of per-backbone leaderboard performance and was not part of the initial design space.
Over 50 variants were generated, with the best reaching rank 10 (AP = 0.9010).

\section{System Design Details}
\label{app:system}

\paragraph{Orchestration.}
The system coordinates multi-agent research using filesystem-based state management across two compute nodes.
Each LLM agent operates as an independent process that reads shared state (leaderboard, experiment logs, configuration registry) and writes new idea proposals as YAML files.
A scheduler daemon assigns ideas to available GPUs based on priority, deduplicating near-identical proposals using configuration fingerprinting (Jaccard similarity $> 0.9$ on the categorical dimensions triggers deduplication).

\paragraph{Self-healing.}
When an experiment fails at runtime, the system captures the full error trace and invokes an LLM to: (i)~diagnose the root cause, (ii)~propose a minimal code patch, and (iii)~re-queue the experiment.
Over the 27-day campaign, 64 auto-fix attempts were made, with 52 succeeding on the first attempt (81\% first-attempt success rate).
Common fixes include: adding NaN guards to gradient computation, reducing batch size for OOM recovery, and handling missing feature files.

\paragraph{Execution pipeline.}
Each experiment follows: feature loading $\to$ temporal encoder training $\to$ validation evaluation $\to$ test evaluation $\to$ metric logging.
Training uses pre-extracted frame features (not end-to-end), enabling rapid iteration ($\sim$3--10 minutes per experiment depending on configuration).
Features are extracted once per backbone using frozen pretrained models and cached on shared storage.

\paragraph{Agent prompt design.}
Each research cycle, the agent receives:
\begin{itemize}[leftmargin=*, itemsep=1pt]
    \item Current leaderboard (top-20 configurations with AP, backbone, encoder, key hyperparameters)
    \item Distribution summary of recent experiments (backbone and encoder frequencies)
    \item Failure log summaries from the last 10 failed experiments
    \item Research rules specifying the configuration schema and diversity requirements
\end{itemize}
The agent outputs 3--5 ideas as structured YAML, each specifying all configuration dimensions in $\Cspace$.

\section{Extended Ablation Tables}
\label{app:ablation}

\begin{figure}[ht]
\centering
\includegraphics[width=0.9\textwidth]{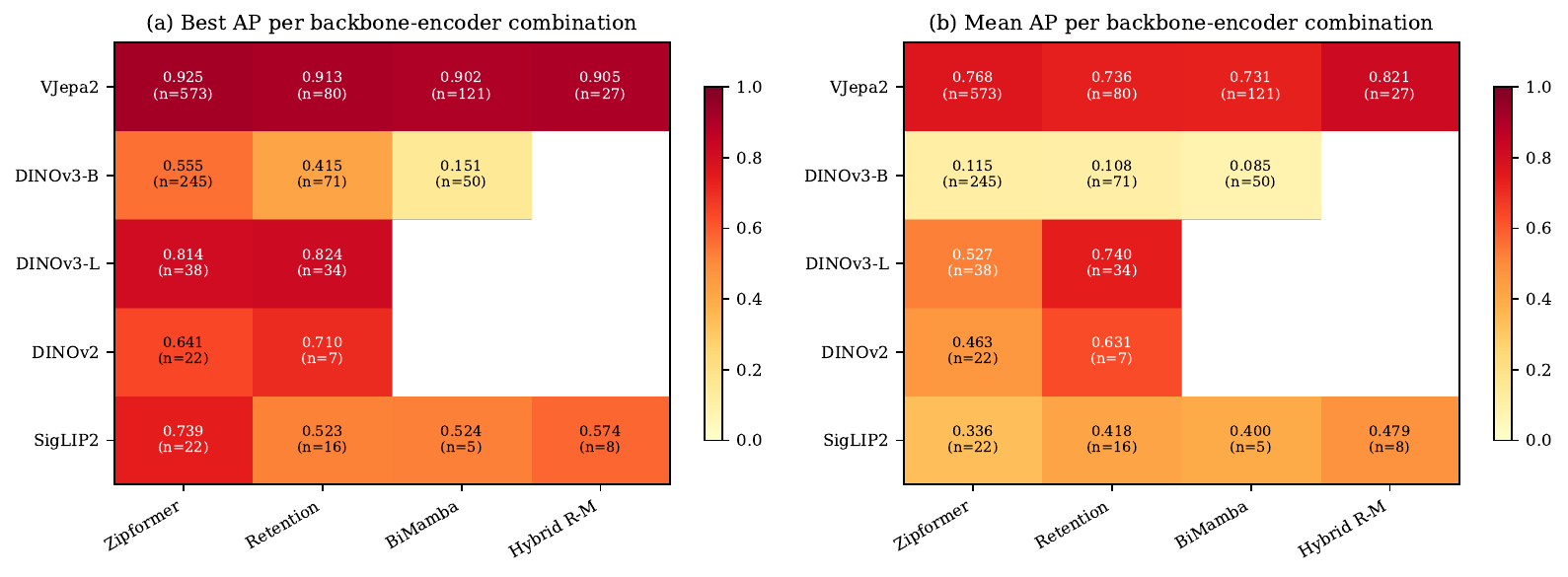}
\caption{Mean AP heatmap for backbone $\times$ encoder combinations. VJepa2 dominates across all encoder types, with Zipformer and Hybrid R-M performing best within VJepa2. DINOv3-B performs poorly across all encoders, suggesting the base-size model lacks capacity for this task.}
\label{fig:heatmap}
\end{figure}

Figure~\ref{fig:heatmap} shows the full backbone $\times$ encoder performance heatmap across all architectural combinations with $\geq$5 experiments.

\section{Temporal Encoder Architecture Details}
\label{app:architectures}

This appendix provides the full mathematical formulations for the four temporal encoder families summarized in Section~\ref{sec:architectures}.

\subsection{Zipformer Temporal Encoder}

\paragraph{BiasNorm.}
Unlike LayerNorm, BiasNorm preserves information about vector magnitude:
\begin{equation}
\mathrm{BiasNorm}(x; b, \gamma) = \frac{x \cdot \exp(\gamma)}{\sqrt{\frac{1}{d}\sum_{j=1}^{d}(x_j - b_j)^2 + \epsilon}}
\end{equation}
where $b \in \R^d$ is a learnable bias and $\gamma \in \R$ is a learnable log-scale parameter.

\paragraph{SwooshR/L activations.}
$\mathrm{SwooshR}(x) = \log(1 + e^{x-1}) - 0.08x - 0.313$; $\mathrm{SwooshL}(x) = \log(1 + e^{x-4}) - 0.08x - 0.035$.

\paragraph{Bypass module.}
Learnable residual interpolation: $\mathrm{Bypass}(x, y; c) = (1 - c_{\mathrm{eff}}) \cdot x + c_{\mathrm{eff}} \cdot y$, where bounds relax linearly over warmup steps.

\paragraph{Zipformer block.}
Three sub-layers with bypass connections:
\begin{align}
h_1 &= \mathrm{Bypass}_{\mathrm{attn}}\big(x,\; x + \mathrm{MHSA}(\mathrm{BiasNorm}(x))\big) \nonumber \\
h_2 &= \mathrm{Bypass}_{\mathrm{conv}}\big(h_1,\; h_1 + \mathrm{Conv1D}(\mathrm{BiasNorm}(h_1))\big) \\
h_3 &= \mathrm{Bypass}_{\mathrm{ff}}\big(h_2,\; h_2 + \mathrm{FFN}(\mathrm{BiasNorm}(h_2))\big) \nonumber
\end{align}

\paragraph{Multi-scale structure.}
$S$ stacks at different temporal resolutions with learned downsampling: $\mathrm{Downsample}(x)_{t'} = \sum_{j=0}^{d-1} \mathrm{softmax}(\alpha)_j \cdot x_{t' d + j}$.

\subsection{RetNet Temporal Encoder}

\paragraph{Single-head retention.}
$\mathrm{Retention}(Q, K, V; \gamma) = (QK^\top / \sqrt{d} \odot D(\gamma)) V$, where $D_{ij}(\gamma) = \gamma^{i-j}$ for $i \geq j$, 0 otherwise.

\paragraph{Multi-scale retention.}
$H$ heads with decay rates $\gamma_h = 1 - 2^{-(5+h)}$:
$\mathrm{MSR}(x) = W_{\mathrm{out}}\, \mathrm{GroupNorm}(\mathrm{Concat}_h[\mathrm{Retention}(Q_h, K_h, V_h; \gamma_h)])$.

\subsection{BiMamba Temporal Encoder}

\paragraph{Selective SSM.}
Discretized via zero-order hold: $\bar{A}_t = \exp(\delta_t A)$, \, $h_t = \bar{A}_t h_{t-1} + \bar{B}_t x_t$, \, $y_t = C(x_t)^{\!\top} h_t + D x_t$.

\paragraph{Bidirectional extension.}
$\mathrm{BiMamba}(x) = \mathrm{LN}(\alpha_{\mathrm{fwd}} \cdot \mathrm{MambaBlock}_{\mathrm{fwd}}(x) + \alpha_{\mathrm{bwd}} \cdot \mathrm{flip}(\mathrm{MambaBlock}_{\mathrm{bwd}}(\mathrm{flip}(x)))) + x$.

\subsection{Hybrid Retention-Mamba}
Layer $2k$: RetentionBlock (quadratic, distance-weighted). Layer $2k{+}1$: BiMambaBlock (linear, sequential state).

\subsection{Attention Pooling}
$z = \sum_{t=1}^{T} \mathrm{softmax}(q^\top W_K x_t / \sqrt{d}) \cdot x_t$, where $q \in \R^d$ is learnable.

\subsection{Focal Loss}
$\mathrm{FL}(p, y; \alpha, \gamma) = -\alpha_y (1 - p_y)^\gamma \log(p_y)$. With $\gamma = 3$, focal loss down-weights well-classified examples, focusing gradient updates on hard cases regardless of class balance.

\section{Notation Table}
\label{app:notation}

\begin{table}[h]
\centering
\caption{Complete notation reference.}
\small
\begin{tabular}{@{}lll@{}}
\toprule
Symbol & Definition & Domain \\
\midrule
$\Cspace$ & Configuration space & $\Cspace_{\mathrm{arch}} \times \Cspace_{\mathrm{loss}} \times \Cspace_{\mathrm{train}} \times \Cspace_{\mathrm{data}}$ \\
$c$ & A specific configuration & $c \in \Cspace$ \\
$f(c)$ & Evaluation function (returns AP) & $f: \Cspace \to [0, 1]$ \\
$y_t$ & Observed AP at step $t$ & $y_t = f(c_t) + \epsilon_t$ \\
$\pi$ & Search policy & $\pi: (\Cspace \times \R)^* \to \Delta(\Cspace)$ \\
$H_t$ & History at step $t$ & $H_t = \{(c_1, y_1), \ldots, (c_t, y_t)\}$ \\
$\APstar(N)$ & Cumulative best AP after $N$ experiments & $\APstar(N) = \max_{i \leq N} y_i$ \\
$r_N$ & Simple regret after $N$ experiments & $r_N = f(c^*) - \max_{i \leq N} f(c_i)$ \\
$H(t)$ & Configuration-space entropy & $H(t) = -\sum_c p_t(c) \log p_t(c)$ \\
$D_{\mathrm{spec}}$ & Agent specialization (JSD) & $D_{\mathrm{spec}} \in [0, \log 2]$ \\
$\iota_t$ & Innovation indicator & $\iota_t = \mathbf{1}[y_t > \APstar(t{-}1)]$ \\
$T$ & Sequence length (temporal) & $T \in \N$ \\
$d$ & Model / feature dimension & $d \in \N$ \\
$\gamma$ & Focal loss focusing parameter & $\gamma \in [0.5, 5.0]$ \\
& \quad or RetNet decay rate & $\gamma \in (0, 1)$ \\
$\alpha$ & Focal loss class weight & $\alpha \in [0, 1]$ \\
$\theta$ & Model parameters & $\theta \in \R^p$ \\
\bottomrule
\end{tabular}
\label{tab:notation}
\end{table}

\section{Research Agent Prompt Examples}
\label{app:prompts}

Below is a representative excerpt of the research agent prompt (anonymized). The full prompt includes the leaderboard, diversity rules, and configuration schema.

\begin{verbatim}
## Current Leaderboard (Top 5)
| Rank | AP    | Backbone | Encoder   | Loss        |
|------|-------|----------|-----------|-------------|
| 1    | 0.924 | VJepa2   | Zipformer | Focal g=3.0 |
| 2    | 0.921 | VJepa2   | Zipformer | Focal g=2.5 |
| 3    | 0.920 | VJepa2   | Zipformer | Focal g=3.0 |
| 4    | 0.916 | VJepa2   | Zipformer | Focal g=3.0 |
| 5    | 0.913 | VJepa2   | Retention | Focal g=2.5 |

## Diversity Budget
- At least 1 idea must use a non-VJepa2 backbone
- At least 1 idea must use a non-Zipformer encoder
- Banned configs: [list of recently failed configs]

## Task
Propose 3-5 ideas as YAML configurations.
\end{verbatim}

The agent responds with structured YAML specifying all dimensions of $\Cspace$:

\begin{verbatim}
- idea_name: "VJepa2 BiMamba with higher dropout"
  backbone: vjepa2_vitl14
  temporal_encoder: bimamba
  loss: focal
  focal_gamma: 3.0
  learning_rate: 3e-4
  weight_decay: 0.05
  batch_size: 64
  seq_len: 20
  epochs: 15
  priority: high
  rationale: "BiMamba offers linear-time
    alternative to Zipformer. Higher dropout
    (0.3) may reduce overfitting seen in
    recent BiMamba runs."
\end{verbatim}

\section{Reproducibility Details}
\label{app:reproducibility}

\paragraph{Hardware.}
$2 \times 8$ NVIDIA H100 80GB GPUs on two nodes connected via shared Lustre filesystem.
Total GPU-hours: 3,227.

\paragraph{Software.}
Python 3.10, PyTorch 2.x, scikit-learn for metrics.
Pre-extracted features from: V-JEPA\,2 \citep{bardes2024vjepa} (1024-dim), DINOv2 \citep{oquab2024dinov2} (768-dim), DINOv3 (768-dim), SigLIP2 \citep{zhai2023siglip} (768-dim).

\paragraph{LLM agents.}
Claude Opus (Anthropic): 819 research cycles, 3--5 base ideas per cycle (expanded by auto-sweep).
Gemini 2.5 Pro (Google): 864 research cycles, 3--5 base ideas per cycle (expanded by auto-sweep).
Temperature: default for both models.
Total LLM API cost: {\raise.17ex\hbox{$\scriptstyle\sim$}}\$2{,}000.

\paragraph{Training details.}
Each experiment trains a temporal encoder (Zipformer, RetNet, BiMamba, or Hybrid) on frozen backbone features.
Training duration: 5--30 epochs ($\sim$3--10 minutes on a single H100).
Optimizer: AdamW.
Learning rate schedule: cosine, linear, or step (configuration-dependent).
Evaluation metric: Average Precision (AP) on the validation split; test AP reported for top configurations.

\paragraph{Dataset.}
Nexar dashcam collision prediction dataset: 1,500 videos (750 collision/near-miss, 750 non-collision), split 80/10/10.
Features extracted at 5 fps, sequence length $T = 20$ frames.
Train/validation/test splits: 1,184\ / 177\ / 139.


\begin{thebibliography}{99}

\bibitem[Akiba et~al.(2019)]{akiba2019optuna}
Akiba, T., Sano, S., Yanase, T., Ohta, T., \& Koyama, M. (2019).
\newblock Optuna: A next-generation hyperparameter optimization framework.
\newblock In \emph{KDD}.

\bibitem[Bergstra et~al.(2011)]{bergstra2011algorithms}
Bergstra, J., Bardenet, R., Bengio, Y., \& K{\'e}gl, B. (2011).
\newblock Algorithms for hyper-parameter optimization.
\newblock In \emph{NeurIPS}.

\bibitem[Bergstra \& Bengio(2012)]{bergstra2012random}
Bergstra, J. \& Bengio, Y. (2012).
\newblock Random search for hyper-parameter optimization.
\newblock \emph{JMLR}, 13:281--305.

\bibitem[Chan et~al.(2024)]{chan2024mlebench}
Chan, J.S., et~al. (2024).
\newblock MLE-bench: Evaluating machine learning agents on machine learning engineering.
\newblock arXiv:2410.07095.

\bibitem[Chen et~al.(2024)]{chen2024evoprompting}
Chen, A., Dohan, D.M., \& So, D. (2024).
\newblock EvoPrompting: Language models for code-level neural architecture search.
\newblock In \emph{NeurIPS 2023}.

\bibitem[Domhan et~al.(2015)]{domhan2015speeding}
Domhan, T., Springenberg, J.T., \& Hutter, F. (2015).
\newblock Speeding up automatic hyperparameter optimization of deep neural networks by extrapolation of learning curves.
\newblock In \emph{IJCAI}.

\bibitem[Falkner et~al.(2018)]{falkner2018bohb}
Falkner, S., Klein, A., \& Hutter, F. (2018).
\newblock BOHB: Robust and efficient hyperparameter optimization at scale.
\newblock In \emph{ICML}. arXiv:1807.01774.


\bibitem[Gu \& Dao(2023)]{gu2023mamba}
Gu, A. \& Dao, T. (2023).
\newblock Mamba: Linear-time sequence modeling with selective state spaces.
\newblock arXiv:2312.00752.

\bibitem[Harris \& Slivkins(2025)]{harris2025explore}
Harris, K. \& Slivkins, A. (2025).
\newblock Should you use your large language model to explore or exploit?
\newblock arXiv:2502.00225.

\bibitem[Hoffmann et~al.(2022)]{hoffmann2022chinchilla}
Hoffmann, J., Borgeaud, S., Mensch, A., et~al. (2022).
\newblock Training compute-optimal large language models.
\newblock In \emph{NeurIPS}. arXiv:2203.15556.

\bibitem[Huang et~al.(2024)]{huang2024mlagentbench}
Huang, Q., Vora, J., Liang, P., \& Leskovec, J. (2024).
\newblock MLAgentBench: Evaluating language agents on machine learning experimentation.
\newblock In \emph{ICML}. arXiv:2310.03302.

\bibitem[Hutter et~al.(2011)]{hutter2011smac}
Hutter, F., Hoos, H.H., \& Leyton-Brown, K. (2011).
\newblock Sequential model-based algorithm configuration.
\newblock In \emph{LION}.

\bibitem[Jiang et~al.(2025)]{jiang2025aide}
Jiang, Z., et~al. (2025).
\newblock AIDE: AI-driven exploration in the space of code.
\newblock arXiv:2502.13138.

\bibitem[Kadra et~al.(2023)]{kadra2023scaling}
Kadra, A., Janowski, M., Wistuba, M., \& Grabocka, J. (2023).
\newblock Scaling laws for hyperparameter optimization.
\newblock In \emph{NeurIPS}. arXiv:2302.00441.

\bibitem[Kaplan et~al.(2020)]{kaplan2020scaling}
Kaplan, J., McCandlish, S., Henighan, T., et~al. (2020).
\newblock Scaling laws for neural language models.
\newblock arXiv:2001.08361.

\bibitem[King et~al.(2004)]{king2004robot}
King, R.D., Whelan, K.E., Jones, F.M., et~al. (2004).
\newblock Functional genomic hypothesis generation and experimentation by a robot scientist.
\newblock \emph{Nature}, 427:247--252.


\bibitem[Lin et~al.(2017)]{lin2017focal}
Lin, T.-Y., Goyal, P., Girshick, R., He, K., \& Doll{\'a}r, P. (2017).
\newblock Focal loss for dense object detection.
\newblock In \emph{ICCV}. arXiv:1708.02002.

\bibitem[Liu et~al.(2019)]{liu2019darts}
Liu, H., Simonyan, K., \& Yang, Y. (2019).
\newblock DARTS: Differentiable architecture search.
\newblock In \emph{ICLR}. arXiv:1806.09055.

\bibitem[Lu et~al.(2024)]{lu2024aiscientist}
Lu, C., Lu, C., Lange, R.T., Foerster, J., Clune, J., \& Ha, D. (2024).
\newblock The AI Scientist: Towards fully automated open-ended scientific discovery.
\newblock arXiv:2408.06292.

\bibitem[Nathani et~al.(2025)]{nathani2025mlgym}
Nathani, D., et~al. (2025).
\newblock MLGym: A new framework and benchmark for advancing AI research agents.
\newblock arXiv:2502.14499.

\bibitem[Moura et~al.(2025)]{nexar2025}
Moura, D.~C., Zhu, S., and Zvitia, O. (2025).
\newblock Nexar dashcam collision prediction dataset and challenge.
\newblock arXiv:2503.03848.

\bibitem[Nexar(2024)]{nexar2024}
Nexar (2024).
\newblock Nexar dashcam collision prediction challenge.
\newblock \url{https://www.kaggle.com/competitions/nexar-collision-prediction}.


\bibitem[Pham et~al.(2018)]{pham2018enas}
Pham, H., Guan, M., Zoph, B., Le, Q., \& Dean, J. (2018).
\newblock Efficient neural architecture search via parameter sharing.
\newblock In \emph{ICML}. arXiv:1802.03268.

\bibitem[Romera-Paredes et~al.(2024)]{romeraparedes2024funsearch}
Romera-Paredes, B., Barekatain, M., Novikov, A., et~al. (2024).
\newblock Mathematical discoveries from program search with large language models.
\newblock \emph{Nature}, 625:468--475.

\bibitem[Schmidgall et~al.(2025)]{schmidgall2025agentlab}
Schmidgall, S., Su, Y., Wang, Z., et~al. (2025).
\newblock Agent Laboratory: Using LLM agents as research assistants.
\newblock arXiv:2501.04227.

\bibitem[Slivkins(2019)]{slivkins2019bandits}
Slivkins, A. (2019).
\newblock Introduction to multi-armed bandits.
\newblock \emph{Foundations and Trends in ML}, 12(1-2):1--286. arXiv:1904.07272.

\bibitem[Snoek et~al.(2012)]{snoek2012practical}
Snoek, J., Larochelle, H., \& Adams, R.P. (2012).
\newblock Practical Bayesian optimization of machine learning algorithms.
\newblock In \emph{NeurIPS}.

\bibitem[Sun et~al.(2023)]{sun2023retnet}
Sun, Y., Dong, L., Huang, S., et~al. (2023).
\newblock Retentive network: A successor to transformer for large language models.
\newblock arXiv:2307.08621.

\bibitem[Tang(2024)]{tang2024rex}
Tang, H. (2024).
\newblock Code repair with LLMs gives an exploration-exploitation tradeoff.
\newblock In \emph{NeurIPS}.

\bibitem[Wijk et~al.(2024)]{wijk2024rebench}
Wijk, H., et~al. (2024).
\newblock RE-Bench: Evaluating frontier AI R\&D capabilities of language model agents against human experts.
\newblock arXiv:2411.15114.

\bibitem[Yang et~al.(2024)]{yang2024opro}
Yang, C., Wang, X., Lu, Y., et~al. (2024).
\newblock Large language models as optimizers.
\newblock In \emph{ICLR}. arXiv:2309.03409.

\bibitem[Yao et~al.(2023)]{yao2023zipformer}
Yao, Z., Guo, L., Yang, X., et~al. (2023).
\newblock Zipformer: A faster and better encoder for automatic speech recognition.
\newblock arXiv:2310.11230.

\bibitem[Ye et~al.(2024)]{ye2024reevo}
Ye, H., et~al. (2024).
\newblock ReEvo: Large language models as hyper-heuristics with reflective evolution.
\newblock In \emph{NeurIPS}.

\bibitem[Zhang et~al.(2025a)]{zhang2025verbalized}
Zhang, J., et~al. (2025).
\newblock Verbalized sampling: How to mitigate mode collapse and unlock LLM diversity.
\newblock arXiv:2510.01171.

\bibitem[Zhang et~al.(2025b)]{zhang2025scaling}
Zhang, P., Zhang, H., Xu, H., et~al. (2025).
\newblock Scaling laws in scientific discovery with AI and robot scientists.
\newblock arXiv:2503.22444.

\bibitem[Bardes et~al.(2024)]{bardes2024vjepa}
Bardes, A., Garrido, Q., Ponce, J., et~al. (2024).
\newblock Revisiting feature prediction for learning visual representations from video.
\newblock arXiv:2404.08471.

\bibitem[Oquab et~al.(2024)]{oquab2024dinov2}
Oquab, M., Darcet, T., Moutakanni, T., et~al. (2024).
\newblock DINOv2: Learning robust visual features without supervision.
\newblock \emph{TMLR}. arXiv:2304.07193.

\bibitem[Zhai et~al.(2023)]{zhai2023siglip}
Zhai, X., Mustafa, B., Kolesnikov, A., \& Beyer, L. (2023).
\newblock Sigmoid loss for language image pre-training.
\newblock In \emph{ICCV}. arXiv:2303.15343.

\bibitem[Sharma et~al.(2024)]{sharma2024sycophancy}
Sharma, M., et~al. (2024).
\newblock Towards understanding sycophancy in language models.
\newblock In \emph{ICLR}. arXiv:2310.13548.


\bibitem[Jimenez et~al.(2024)]{jimenez2024swebench}
Jimenez, C.E., Yang, J., Wettig, A., et~al. (2024).
\newblock SWE-bench: Can language models resolve real-world GitHub issues?
\newblock In \emph{ICLR}. arXiv:2310.06770.


\end{thebibliography}
\end{document}